\documentclass[letterpaper, 10 pt, conference]{ieeeconf}  

\usepackage[whole]{bxcjkjatype} 

\IEEEoverridecommandlockouts                              

\usepackage{graphics} 
\usepackage{epsfig} 
\usepackage{mathptmx} 
\usepackage{times} 
\usepackage{amsmath} 
\usepackage{amssymb}  

\usepackage{multirow}
\usepackage[table,xcdraw]{xcolor}
\usepackage{threeparttable}
\usepackage{eucal}
\usepackage{balance}
\usepackage{mathrsfs}  
\usepackage{comment}
\usepackage[noend]{algpseudocode}
\usepackage{float}
\usepackage{textcomp}
\usepackage{mathcomp}
\usepackage{bm}
\usepackage{cite} 
\usepackage{booktabs}
\usepackage{url}

\usepackage{hyperref}

\def\ie{\textit{i.e}\onedot}

\makeatletter
\let\MYcaption\@makecaption
\makeatother

\usepackage[font=footnotesize]{subcaption}

\makeatletter
\let\@makecaption\MYcaption
\makeatother

\usepackage{subcaption}

\algnewcommand{\Break}{\textbf{break}}

\newcommand{\myvec}[1]{\mathbf{#1}}
\newcommand{\myarray}[1]{\textbf{\textit{#1}}}
\newcommand{\myset}[1]{\mathcal{#1}}

\newcommand{\mytag}[1]{\texttt{#1}}

\def\ie{{\it i.e.}}

\usepackage[colorinlistoftodos]{todonotes}

\title{\LARGE \bf
GSplatVNM: Point-of-View Synthesis\\for Visual Navigation Models Using Gaussian Splatting
}

\author{Kohei Honda$^{1, 2}$, Takeshi Ishita$^{1}$, Yasuhiro Yoshimura$^{1}$, Ryo Yonetani$^{1}$
\thanks{$^{1}$CyberAgent AI Lab, Tokyo, Japan, {\tt\small \{honda\_kohei, ishita\_takeshi, yoshimura\_yasuhiro, yonetani\_ryo\}@cyberagent.co.jp}}%
\thanks{$^{2}$The Department of Mechanical Systems Engineering, Nagoya University, Aichi, Japan, {\tt\small honda.kohei.b0@s.mail.nagoya-u.ac.jp}}%
}

\begin{document}

\maketitle

\thispagestyle{empty}
\pagestyle{empty}


\begin{abstract}
This paper presents a novel approach to image-goal navigation by integrating 3D Gaussian Splatting (3DGS) with Visual Navigation Models (VNMs), a method we refer to as GSplatVNM. 
VNMs offer a promising paradigm for image-goal navigation by guiding a robot through a sequence of point-of-view images without requiring metrical localization or environment-specific training. 
However, constructing a dense and traversable sequence of target viewpoints from start to goal remains a central challenge, particularly when the available image database is sparse. 
To address these challenges, we propose a 3DGS-based viewpoint synthesis framework for VNMs that synthesizes intermediate viewpoints to seamlessly bridge gaps in sparse data while significantly reducing storage overhead. Experimental results in a photorealistic simulator demonstrate that our approach not only enhances navigation efficiency but also exhibits robustness under varying levels of image database sparsity.

\end{abstract}


\section{INTRODUCTION}
\label{sec:introduction}

\looseness=-1 
Efficient robot navigation relies on the availability of sufficient environmental information; however, the associated data collection costs cannot always be justified. For example, imagine an exhibition hall or retail store. The appearance and structure of such environments may change day by day, necessitating continuous monitoring and updates. Moreover, regular site surveys can only be conducted during the limited hours when the facility is closed. Therefore, reducing overall data collection costs is essential for deploying robots in these settings.

Vision-based navigation offers a promising solution. 
In particular, recent advances in image-to-waypoints Visual Navigation Models (VNMs)~\cite{shah2023gnm,shah2023vint,sridhar2024nomad} have demonstrated zero-shot navigation capabilities without precise self-localization using dense point clouds~\cite{thoma2019mapping,kwon2023renderable}, thereby reducing the need for extensive data collection compared to conventional localization-based navigation frameworks. 
Nevertheless, these approaches have not completely eliminated the need for data collection; VNMs plan a path based on the robot's point-of-view images of the goal, and a continuous sequence of such images—from the start to the goal—is necessary for long-range navigation. Existing methods construct an Image-based Topological Graph (ITG) to describe the traversability between randomly sampled images in the environment~\cite{fraundorfer2007topological}, which necessitates collecting a sufficient number of images to ensure robust navigation performance.

\begin{figure}[t]
    \centering
    \includegraphics[width=\linewidth]{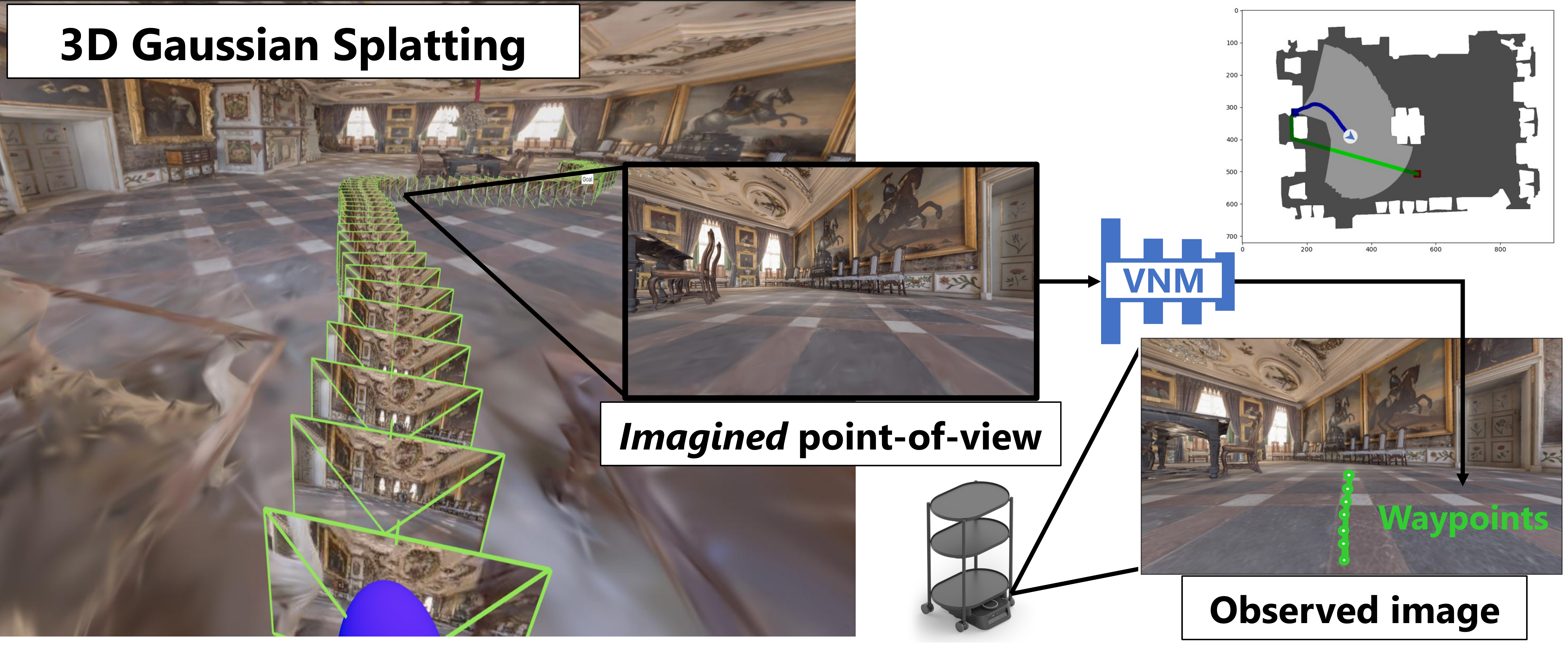}
    \caption{GSplatVNM employs 3DGS as a compact and renderable environment representation for the VNM, which \emph{imagines} future point-of-view images to efficiently navigate the robot from the given start to the goal.}
    \label{fig:eye_catch}
\end{figure}

\looseness=-1 
In this work, we propose \emph{GSplatVNM}, a new vision-based navigation framework that requires reduced data collection. 
As illustrated in Fig.~\ref{fig:eye_catch}, we integrate VNMs with 3D Gaussian splatting (3DGS)~\cite{kerbl20233d} as a compact environment representation. 
3DGS is a neural model that enables high-quality 3D reconstruction of the environment from a pre-collected image database (DB) and can further synthesize novel images for arbitrary viewpoints not present in the original database. We leverage these capabilities of 3DGS to localize the start and goal poses from a given pair of images and connect them by synthesizing a continuous sequence of traversable point-of-view images. These images are then used by the VNM to control the robot.

We validate the proposed GSplatVNM framework in photorealistic simulation environments~\cite{xiazamirhe2018gibsonenv,habitat19iccv}. 
Our experiments demonstrate that GSplatVNM outperforms conventional ITG-based methods~\cite{sridhar2024nomad, shah2023lm} in terms of navigation efficiency and robustness, particularly when using sparse image databases. 
Notably, GSplatVNM can even navigate to a point-of-view that has been seen but not visited, a task that has proven difficult for ITG-based methods. 
These results demonstrate that GSplatVNM can significantly reduce data collection costs while maintaining a continuous set of feasible target point-of-view images.
\section{RELATED WORK}
\label{sec:related_work}

\looseness=-1 
Since this paper proposes a novel environment representation for VNMs as an alternative to ITG, we first provide an overview of ITG, which is a commonly used environment representation for visual navigation. We then review recent studies that use neural rendering models as a prior map for visual navigation, which are closely related to our proposed method.

\subsection{ITG-based Visual Navigation}

As shown in Fig.~\ref{fig:system_overview}, ITG is a graph representation of the environment, where each node represents a point-of-view image and edges encode traversability~\cite{fraundorfer2007topological}. 
Unlike point cloud maps, ITG does not contain absolute position information; it is defined by the images and the relative relationships between them. 
Typically, the robot localizes itself within the ITG and navigates by following a path computed using graph search algorithms~\cite{savinov2018semi}. 
Because the robot only needs to estimate its current node in the ITG, high-precision localization and mapping are not required, which enhances portability and scalability. 
Furthermore, ITG-based visual navigation has demonstrated strong generalization performance~\cite{savinov2018semi} to unseen environments compared to \emph{pure} learning-based approaches that do not leverage prior knowledge of the environment~\cite{habitat19iccv, wijmansdd}.

\subsection{Construction of Environment-Covering ITG}

Although ITG-based environment representations are a promising approach for visual navigation, they have a critical limitation: while the robot moves continuously in space, the ITG is represented as a discrete graph. 
As a result, the ITG alone may provide a sparse and insufficient spatial representation, potentially degrading navigation performance. 
For example, the start and goal images may not always be present in the ITG, and suboptimal connectivity estimates between nodes can lead the robot to take unnecessarily long paths. 
Therefore, building an ITG that efficiently covers the environment is a challenging task.

One straightforward approach to constructing an ITG is to physically collect images through robot exploration and then link nodes that are known to be physically traversable~\cite{shah2022viking, shah2022rapid}. 
However, this approach requires extensive sequences of images across the environment. 
Other methods relax this physical requirement by estimating traversability between nodes based on image similarity, thereby allowing edges between nodes that have not been observed through direct physical transitions~\cite{savinov2018semi, shah2023lm, shah2021ving}. 
Although this approach reduces the need for a continuous image sequence, it often struggles in scenarios where the collected image database is spatially sparse, making it difficult to generate a feasible and efficient view sequence to reach the goal. To mitigate this issue, some studies have proposed online updating and expansion of the ITG using simultaneous localization and mapping techniques~\cite{wiyatno2022lifelong, chaplot2020neural} as well as neural image generation models~\cite{shah2023gnm, cui2024frontier}. However, as the number of nodes increases, the boundary between ITG and metric maps becomes blurred, and the computational and storage costs increase significantly.

\subsection{Visual Navigation with Neural Rendering Models}

To overcome the limitations of ITGs, a promising direction is to use neural rendering models as a prior map for visual navigation. 
Some studies generate target point-of-view images online using world models~\cite{koh2021pathdreamer, bar2025navigation}, but these methods require significant computation and lack geometric information about the environment. 
In contrast, Neural Radiance Fields (NeRF)~\cite{mildenhall2021nerf} and 3DGS~\cite{chen2024splat} provide geometric information and can quickly generate novel point-of-view images, making them well-suited for navigation tasks. In particular, 3DGS relies on lightweight 3D Gaussians as its physical representation, which have been used as a geometric prior map for visual navigation~\cite{chen2024splat, lei2025gaussnav, chen2024safer, adamkiewicz2022vision}. 

While such recent studies~\cite{chen2024splat, lei2025gaussnav, chen2024safer, adamkiewicz2022vision} also use 3DGS for navigation, their approaches typically require online and presize self-localization within the 3DGS map. In contrast, our method uses 3DGS as an offline environment model solely to synthesize a sequence of target viewpoints. A powerful, pre-trained VNM then follows these images without needing to explicitly track its pose against the map. Our core contribution is therefore the integration of 3DGS as a viewpoint generator to guide a localization-free policy, rather than using it as a map for online localization.

\section{3DGS AS ENVIRONMENT REPRESENTATION}

\subsection{Representation by 3D Gaussians}
3DGS is an advanced technique for the reconstruction and rendering of environments. 
Instead of using traditional mesh or voxel-based approaches, environments are encoded as a set of 3D Gaussians, each associated with color, opacity, and covariance parameters. 
An environment is represented by 3D Gaussians, denoted as 
$\{(\myvec{\mu}_i, \Sigma_i, \myvec{c}_i, \alpha_i)\}_{i=1}^{N}$,  
where $N$ is the number of Gaussians, $\myvec{\mu}_i \in \mathbb{R}^3$ is the mean, $\Sigma_i \in \mathbb{R}^{3 \times 3}$ is the covariance matrix, $\myvec{c}_i \in \mathbb{R}^3$ is the color, and $\alpha_i \in [0, 1]$ is the opacity of the $i$-th Gaussian. Each Gaussian can be expressed as a continuous function over the 3D position space $\myvec{x} \in \mathbb{R}^3$ as follows:
$G_i(\myvec{x}) = \exp\left(-\frac{1}{2}(\myvec{x} - \myvec{\mu}_i)^{\top} \Sigma_i^{-1} (\myvec{x} - \myvec{\mu}_i)\right).$ 
To render a view from a given camera pose, we \emph{splat} the Gaussians onto the 2D image plane and determine the pixel colors using $\myvec{c}_i$ and $\alpha_i$. By training the parameters of 3DGS to minimize a reprojection error across multiple views, we obtain a compact and renderable representation of the environment.

\subsection{Distance Estimation between Robot and 3DGS}
3DGS can not only render novel views but also be employed to represent the geometry of the environment~\cite{chen2024splat,chen2024safer,lei2025gaussnav}. 
In this work, following the approach proposed in \cite{chen2024safer}, we convert the 3DGS into a set of ellipsoids $\myset{E}_i$ to estimate the distance between the robot and the environment, where each ellipsoid $\myset{E}_i$ is defined as
$\myset{E}_i = \{\myvec{x} \in \mathbb{R}^3 \mid (\myvec{x} - \myvec{\mu}_i)^{\top} \Sigma_i^{-1} (\myvec{x} - \myvec{\mu}_i) \leq 1\}.$
Thus, the ellipsoid-to-sphere distance between the $i$-th ellipsoid and the robot centroid 
$\hat{\myvec{p}}=\myarray{F}\myarray{R}^{\top} (\myvec{p} - \myvec{\mu}_i)$
can be formulated as the following quadratic program:
\begin{align}
  d_i (\myvec{p}) = \min_{\myvec{x} \in \mathbb{R}^3} \|\myvec{x} - \hat{\myvec{p}}\|^2, \quad \text{s.t.} \;\myvec{x}^{\top} \myarray{S}^{-2} \myvec{x} \leq 1, \label{eq:distance}
\end{align}
where $\myvec{p} \in \mathbb{R}^3$ is the robot 3D position approximated by a sphere,  $\myarray{F}$ is a reflection matrix (a diagonal matrix with entries of either 1 or $-1$ used to reflect points across the primary planes), $\myarray{R}$ is the rotation matrix of the ellipsoid, and $\myarray{S}$ is a scaling matrix determined by the confidence of the Gaussian, which serves as a hyperparameter.

\section{VISUAL NAVIGATION WITH 3DGS}
\label{sec:method}

\begin{figure*}[t]
    \centering
    \includegraphics[width=\linewidth]{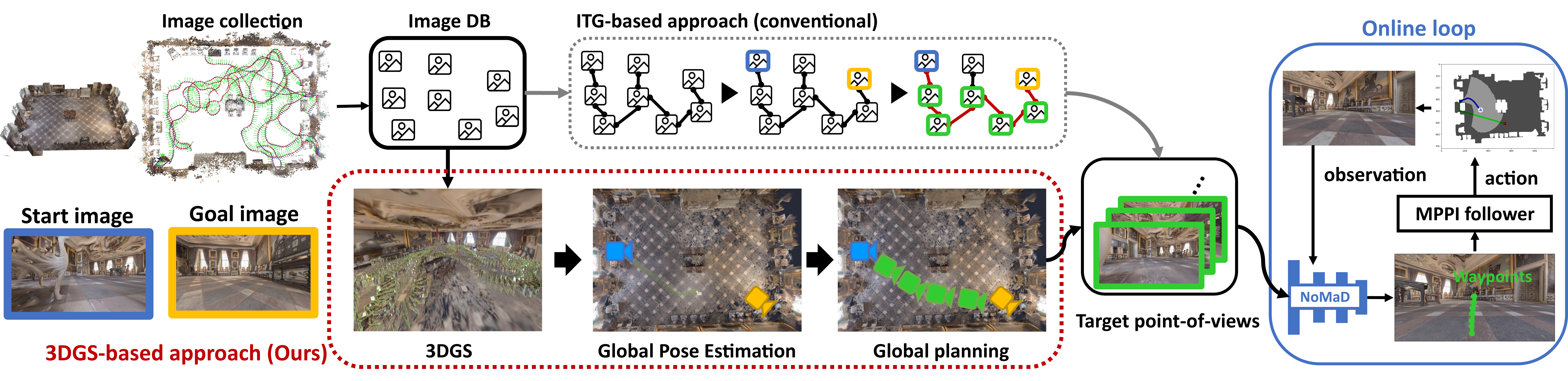}
    \caption{Overview of the proposed GSplatVNM. 
    In a conventional ITG-based approach, the environment is represented by ITG, and the target point-of-views given to the VNM (NoMaD~\cite{sridhar2024nomad}) are limited to a set of collected image DB. 
    On the other hand, GSplatVNM represents the environment as 3DGS to synthesize dense and traversable point-of-view images from the start to the goal, thereby guiding NoMaD.
    As a result, GSplatVNM can achieve efficient image-goal navigation by seamlessly bridging gaps in the image DB, even when the image DB is sparse.
    }
    \label{fig:system_overview}
\end{figure*}

Our aim is to achieve efficient image-goal navigation with VNMs using a compact and spatially aware representation of the navigation environment. 
Our proposed framework uses 3DGS as a prior map for the VNM to reduce the number of images that need to be collected beforehand.

\subsection{Overview of the Proposed Framework}
\looseness=-1 
Figure~\ref{fig:system_overview} shows an overview of our proposed navigation framework integrating 3DGS with a VNM. 
Given the start and goal images, we first estimate the robot's start and goal poses in 3DGS and then plan a global trajectory within the 3DGS representation. 
We subsequently render target point-of-view images along the planned trajectory to condition the VNM, \ie, NoMaD~\cite{sridhar2024nomad}, which is a state-of-the-art VNM. 
During navigation, NoMaD generates spatial waypoints from the observation images and the synthesized target point-of-view images, and the robot follows these waypoints using Model Predictive Path Integral control~(MPPI)~\cite{williams2018information}.

\subsection{Start and Goal Pose Estimation in 3DGS}

In the proposed framework, we first estimate the start and goal poses in 3DGS from the given start and goal images. We assume that the robot moves on a 2D plane; hence, the robot's pose is represented as 
$\myvec{q} = (x, y, \theta_{\rm{yaw}}) \in \mathrm{SE}(2),$
where \(x\) and \(y\) denote the 2D position and \(\theta_{\rm{yaw}}\) the yaw angle. 
Although some studies have addressed accurate and online pose estimation in 3DGS~\cite{chen2024splat}, we require only a one-shot, rough estimation of the global pose.

In this work, we optimize the global start and goal poses, \(\myvec{q}_{\rm{start}}\) and \(\myvec{q}_{\rm{goal}}\), by minimizing the following loss function \(\mathcal{L}\):
\begin{align}
& \mathcal{L}(\myvec{q}_*) = \mathcal{L}_{\text{img}}(I_{*}, I_{\text{rendered}}(\myvec{q}_*)) + \mathbf{1}_{\text{collision}}[d_{\text{min}}(\myvec{q}_*) \leq r ], \label{eq:loss} \\ 
& d_{\text{min}}(\myvec{q}_*) = \min_{i \in \{0, \dots, N-1\} } d_i(\myvec{q}_*), \quad * \in \{\rm{start}, \rm{goal}\}, \nonumber
\end{align}
where \(I_{*}\) is the given start or goal image, \(I_{\text{rendered}}(\myvec{q}_*)\) is the image rendered from 3DGS at pose \(\myvec{q}_*\), and \(r\) is the radius of the robot. 
The first term, \(\mathcal{L}_{\text{img}}\), in (\ref{eq:loss}) is an image similarity metric between the given and rendered images from 3DGS at the pose \(\myvec{q}_*\). Specifically, we use the Learned Perceptual Image Patch Similarity (LPIPS) metric~\cite{zhang2018unreasonable}, which is computed from the feature maps of AlexNet~\cite{krizhevsky2012imagenet} and ranges from 0 to 1. 
The second term is a collision penalty to avoid the infeasibility of global planning. Here, \(\mathbf{1}_{\text{collision}}\) denotes an indicator function, and \(d_i\) is the distance between the robot and the \(i\)-th ellipsoid-approximated 3DGS defined in (\ref{eq:distance}).

To render the image and calculate the distance, in addition to the robot's 2D pose \(\myvec{q}_*\), the \(z\) position \(z_*\), roll \(\theta^*_{\rm{roll}}\), and pitch \(\theta^*_{\rm{pitch}}\) angles are required. We assume that the image DB was captured with fixed \((z, \theta_{\rm{roll}}, \theta_{\rm{pitch}})\) values, and we use the mean of those values obtained from the estimated camera poses during the training of 3DGS.

Although the loss function (\ref{eq:loss}) is theoretically differentiable, its complexity may lead to convergence at local optima. Given that the optimization is over a three-dimensional space, we employ a black-box optimization method known as the Tree-structured Parzen Estimator~\cite{bergstra2011algorithms}. The entire optimization process is implemented using Optuna~\cite{optuna_2019}.

\subsection{Global Trajectory Planning in 3DGS}
\label{sec:global_planning}
We then plan a traversable global trajectory in 3DGS from the estimated start to goal poses, \(\myvec{q}_{\rm{start}}\) and \(\myvec{q}_{\rm{goal}}\). In this work, considering both the guarantee of optimality and ease of implementation, we plan a global path 
$\{(x_{\rm{start}}, y_{\rm{start}}), (x^0, y^0), \dots, (x_{M-1}, y_{M-1}), (x_{\rm{goal}}, y_{\rm{goal}})\}$
using A* search~\cite{hart1968formal} without considering yaw angles, and then assigning yaw angles along the path. That is, we compute the traversable trajectory \(\myset{T}\) as follows:
\begin{align}
    & \myset{T} = \{ \myvec{q}_{\rm{start}}, (x_0, y_0, \theta_{\rm{yaw}}^0), \dots, (x_{M-1}, y_{M-1}, \theta_{\rm{yaw}}^{M-1}), \myvec{q}_{\rm{goal}}\}, \label{eq:path} \\
    & \{(x_0, y_0), \dots, (x_{M-1}, y_{M-1})\} = \text{A*}(\myvec{q}_{\rm{start}}, \myvec{q}_{\rm{goal}}), \nonumber \\
    & \theta_{\rm{yaw}}^i = \text{atan2}(y_{i+1} - y_i, x_{i+1} - x_i), \quad i \in \{0, \dots, M-1\}. \nonumber
\end{align}
Here, \(M\) denotes the number of poses in the trajectory. A* search considers collisions between the robot and the 3DGS as well as the loss function (\ref{eq:loss}). Although the above method guarantees optimality in terms of distance from the start to the goal position, the yaw angles do not account for the robot's dynamics or NoMaD's tracking capability, which may result in trajectories that are not always trackable. We also experimented with Hybrid A* search~\cite{dolgov2008practical}, which can account for the differential two-wheel model, but we found that its performance was not significantly different from (\ref{eq:path}). Additionally, since 3DGS operates on a different scale than the physical world, it is challenging to accurately account for kinodynamics.

\looseness=-1 
After the global planning, we render a set of the target point-of-views $\myset{{V}}_r = \{I_0^r(\myvec{q}_0), \dots, I_{M-1}^r(\myvec{q}_{M-1}), I_{\rm{goal}}\}$ along the planned trajectory $\myset{T}$ using 3DGS.
These rendered images are then used to condition the NoMaD policy to generate spatial waypoints.

\subsection{Zero-shot Local Planning and Control with NoMaD}
\label{sec:nomad}
NoMaD~\cite{sridhar2024nomad} is a visual subgoal-conditioned policy that generates spatial waypoints from a sequence of observation images at time (t), $\myset{O}_{\rm{obs}} = \{I_{t-p}, \dots, I_{t} \}$ (with p=3 in NoMaD), as well as from the target subgoal point-of-view image $I_{\rm{target}}$.
NoMaD consists of three networks:
\begin{itemize}
\item A subgoal image-conditioned vision encoder, $\myvec{c}_t = f_{\rm{enc}}(\myset{O}_{\rm{obs}}, I_{\rm{target}})$, that extracts context features from the observation $\myset{O}_{\rm{obs}}$ and target subgoal image $I_{\rm{target}}$.
\item A diffusion model~\cite{ho2020denoising}-based policy, $\pi(\myvec{c}_t)$, that generates a set of waypoints $\myset{W}_t = \{\myvec{w}_0, \dots, \myvec{w}_{L-1} \}$ (with L=8 in NoMaD) from the encoded features $\myvec{c}_t$, where each waypoint represents a 2D physically-scaled position in the robot's coordinate frame.
\item A distance-estimation network, $\phi(\myvec{c}_t)$, that predicts the physical distance between the current image position and the target image position.
\end{itemize}
These networks are trained in a supervised manner on large, heterogeneous datasets collected across a diverse set of environments and robotic platforms over 100 hours, resulting in zero-shot navigation capability in unseen environments and on unseen robot setups~\cite{sridhar2024nomad}.

In our navigation scheme, after synthesizing a set of target point-of-view images toward the goal, $\myset{{V}}_r$, we first estimate the nearest image $I_{t}^r \in \myset{V}_r$ to the current observation. This estimation is performed using the distance-estimation network, $\phi$, which is conditioned on both the current observation $\myset{O}_{\rm{obs}}$ and the target images $I_{i}^r \in \myset{{V}}_r$ (with $I_0^r = I_{\rm{start}}$). Then, using the subsequent image in the sequence, $I_{t+1}^r$, as the next subgoal, we generate point-of-view-conditioned waypoints $\myset{W}_t$ according to the policy $\pi(\myvec{c}_t)$, where $\myvec{c}_t = f_{\rm{enc}}(\myset{O}_{\rm{obs}}, I_{t+1}^r)$. 
The robot follows these waypoints using MPPI~\cite{williams2018information}, a sampling-based model predictive control method that minimizes future deviations from the waypoints while taking into account the robot's differential two-wheel dynamics.
\section{EXPERIMENTS}
\label{sec:experiment}

We comprehensively evaluate the proposed GSplatVNM on photorealistic 3D indoor environments using the AI Habitat simulator platform~\cite{habitat19iccv}. 
In our experiments, we compare the proposed method with conventional methods in terms of success rate, path efficiency, and robustness with respect to the number of pre-collected images in the image DB.

\subsection{Simulation Setup}
\subsubsection{Robot Setup}
We simulate a circular wheeled robot (radius: 0.5 m) that navigates the environment using the Habitat simulator API, with state updates every 0.5 seconds. The action space comprises continuous linear and angular velocities, with a maximum linear velocity of 0.25 m/s and a maximum angular velocity of 0.5 rad/s. The robot is equipped with a monocular camera mounted at a height of 0.4 m from the ground. The camera has a resolution of 1280$\times$720 pixels and a field of view of 120 degrees.

\subsubsection{Simulation Environments}
To evaluate performance across different environment sizes, we conduct experiments in three indoor environments: \mytag{Greigsville} (small) and \mytag{Ribera} (medium) from the Gibson dataset~\cite{xiazamirhe2018gibsonenv}, and \mytag{skokloster-castle} (large) from the Habitat Test dataset\footnote{\url{https://huggingface.co/datasets/ai-habitat/habitat_test_scenes}}. Their sizes are 43.6, 50.6, and approximately 190 $\rm{m}^2$, respectively.

\subsection{Experiment Setup}

\subsubsection{Image-Goal Navigation Task}
The task is image-goal navigation, where the robot must move from a start pose to a goal position at which the given goal image is visible, while remaining within the traversable area provided by the simulator. A trial is considered successful if the robot reaches the goal within a specified distance (0.5 m for \mytag{Greigsville} and \mytag{Ribera}, and 1.0 m for \mytag{skokloster-castle}) within 500 steps (\ie, 250 s).

In our experiments, we assume that the robot is equipped with a collision avoidance system independent of NoMaD. Consequently, the simulator restricts the robot from leaving the traversable area, and collision avoidance performance is not evaluated\footnote{NoMaD itself has limitations in collision avoidance, which we leave for future work.}.

\subsubsection{Pre-Collection of Image DB}
For the image-goal navigation task, the robot utilizes a pre-collected image DB as prior knowledge of the environment. 
To evaluate robustness with respect to the size of the image DB, we prepare multiple image DBs for each environment. 
Specifically, we collect $\{300, 500, 1000\}$ images for \mytag{Greigsville}, $\{500, 800, 1000\}$ images for \mytag{Ribera}, and $\{300, 1000, 2000, 3000\}$ images for \mytag{skokloster-castle}. 
Before the navigation task, the robot explores the environment using an exploration policy based on NoMaD~\cite{sridhar2024nomad} and collects images at regular intervals of 0.5 s until these numbers of images are collected.

\subsubsection{Evaluation Metrics}
\begin{figure*}[t]
    \centering
    \subfloat[\mytag{Greigsville} (small)\label{fig:spl_greigsville}]{%
        \includegraphics[width=0.32\linewidth]{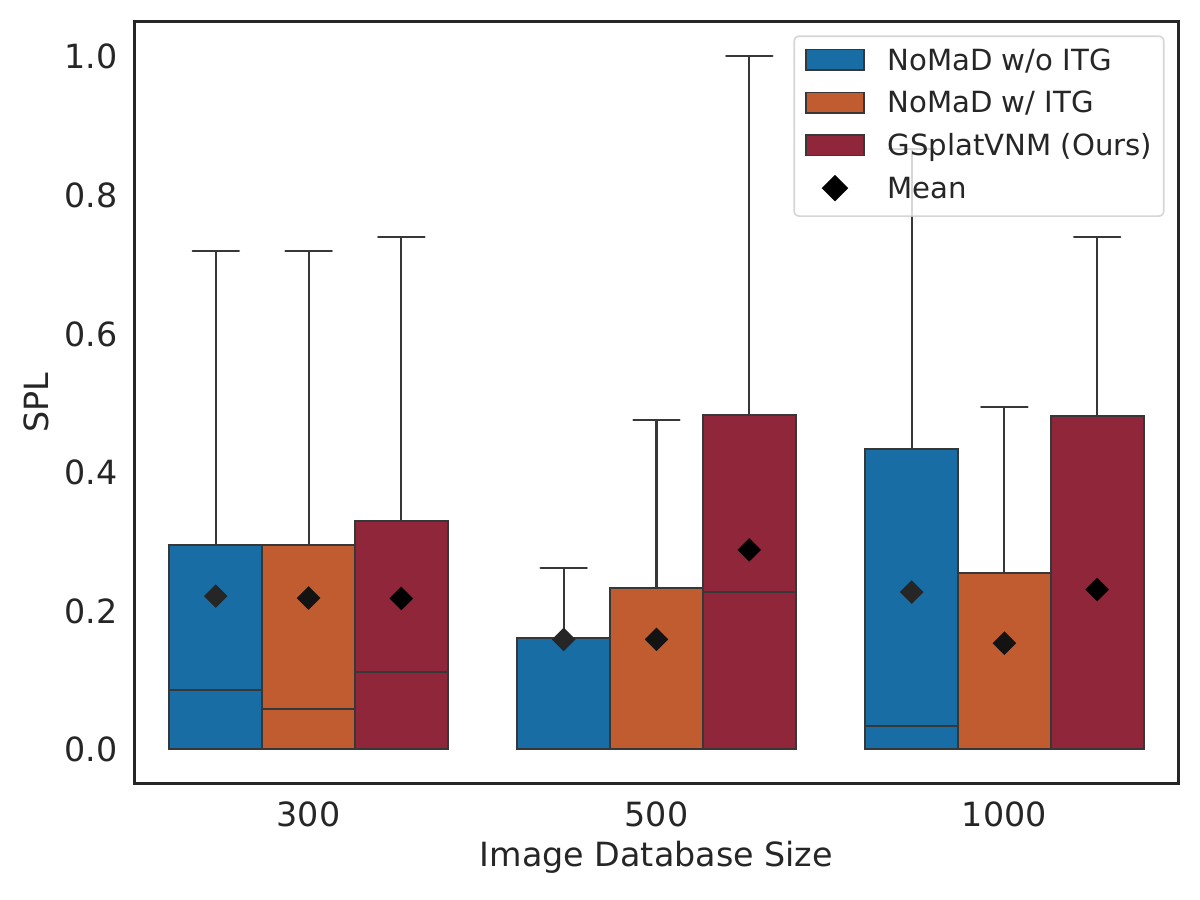}
    }
    \subfloat[\mytag{Ribera} (medium)\label{fig:spl_ribera}]{%
        \includegraphics[width=0.32\linewidth]{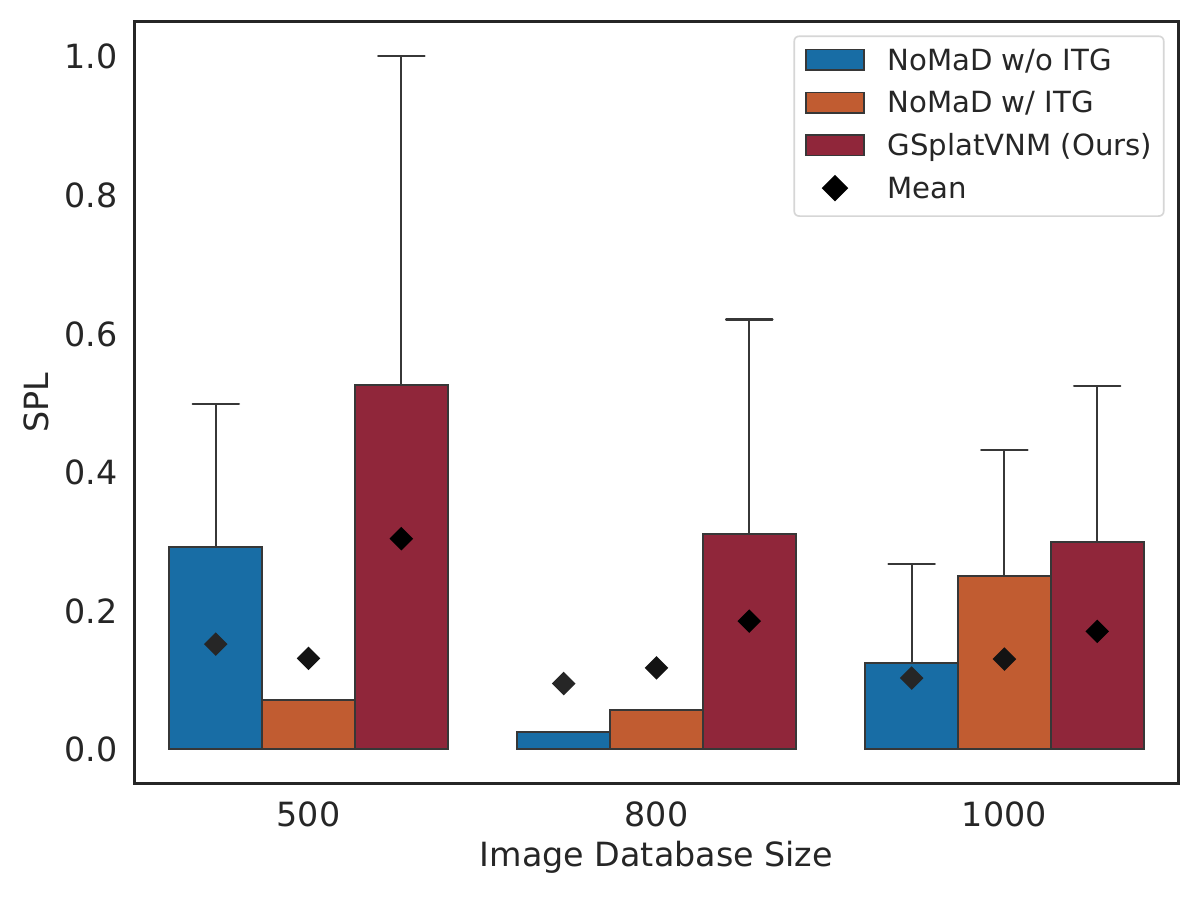}
    }
    \subfloat[\mytag{skokloster-castle} (large)\label{fig:spl_skokloster_castle}]{%
        \includegraphics[width=0.32\linewidth]{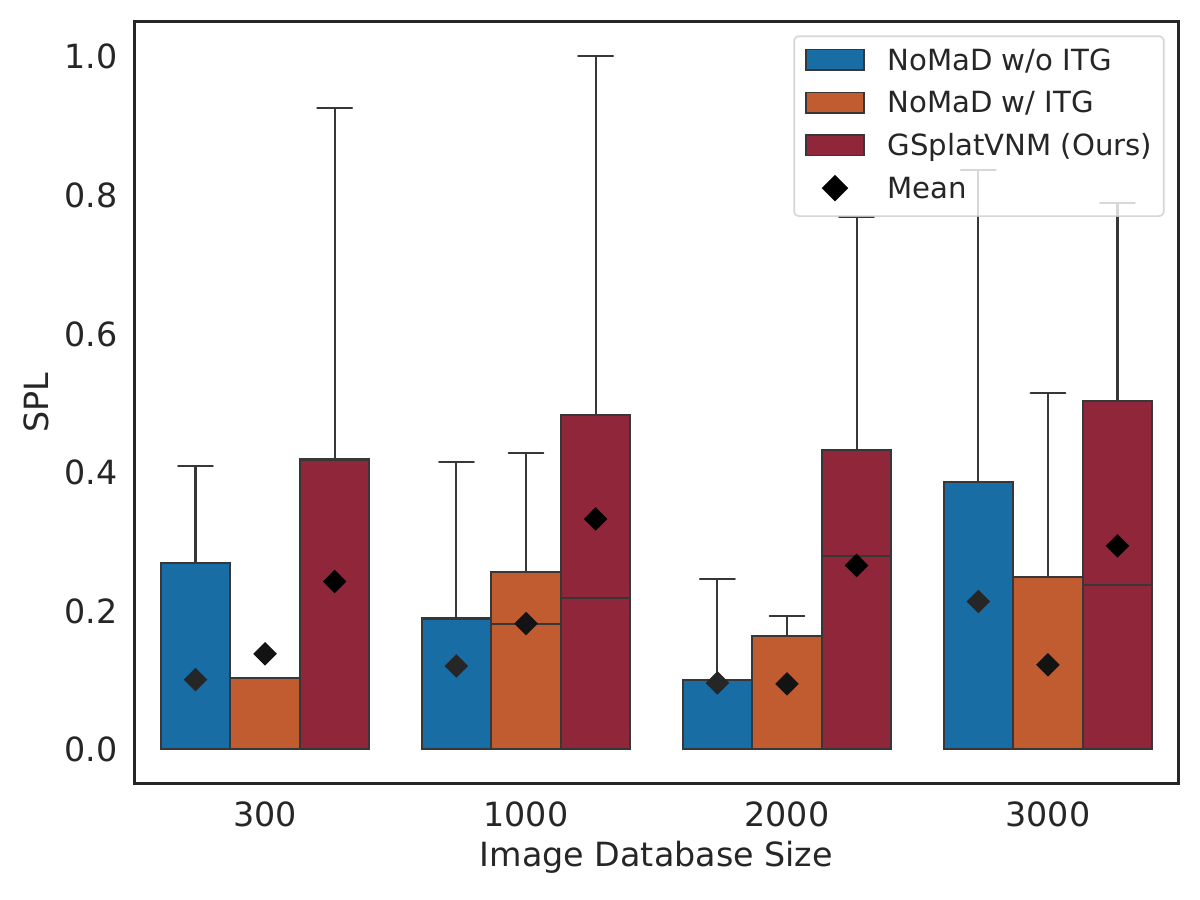}
    }
    \caption{Comparison of the SPL on each environment.}
    \label{fig:spl}
\end{figure*}

We evaluate navigation performance using the Success weighted by Path Length (SPL) metric~\cite{anderson2018evaluation}, which is commonly used to assess the efficiency of navigation tasks. 
SPL is defined as the ratio of the shortest path length to the path length taken by the robot, weighted by the success rate. The definition of SPL is as follows:
\begin{align}
    \text{SPL} = \frac{1}{N_{\rm{trials}}} \sum_{i=1}^{N_{\rm{trials}}} \mathbf{1}_{\rm{success}} \frac{L_i}{\max(L_i, L_{\text{min}})},
\end{align}
where $N_{\rm{trials}}$ is the number of trials, $L_i$ is the path length of the $i$-th trial, $L_{\text{min}}$ is the shortest path length, and $\mathbf{1}_{\rm{success}}$ is an indicator function that returns 1 if the robot reaches the goal. The SPL ranges from 0 to 1, with 1 indicating that the robot reached the goal via the shortest possible path. The shortest path length is computed using the Dijkstra algorithm on the traversable area map provided by the simulator.

\subsection{Implementation Details}

\subsubsection{3DGS Construction}

We construct the 3DGS for each image DB using the Splatfacto implementation in Nerfstudio~\cite{tancik2023nerfstudio}, with the initial Gaussian means provided by a point cloud generated by COLMAP~\cite{schoenberger2016sfm}. 
Note that the robot's radius used in (\ref{eq:distance}) is set to 0.01 in the 3DGS due to the scale difference between the 3DGS and the actual environment.

\subsubsection{Baseline Methods}

To evaluate the effectiveness of our 3DGS-based target point-of-view generation for NoMaD, we compare our approach with the following baseline methods:
\begin{itemize}
    \item \textbf{NoMaD w/o ITG}: NoMaD is conditioned solely on physically reachable point-of-view images~\cite{sridhar2024nomad}. Specifically, the sequence of images traversed (either in order or in reverse) during the image collection, from the estimated start image to the goal image, is used as the sequence of target point-of-views.
    \item \textbf{NoMaD w/ ITG}: NoMaD is conditioned on the shortest target point-of-view sequence computed on a pre-constructed ITG using the Dijkstra algorithm. The ITG is built from an image DB by connecting nodes based on image similarity; two nodes are connected if the estimated distance $\phi$ between them is within a specified threshold (5 m)~\cite{shah2023lm} (see Section~\ref{sec:nomad} for details).
\end{itemize}
In both cases, the start and goal nodes are determined from the image DB as those with the smallest estimated distance $\phi$ to the given start and goal images.

Our goal is to specifically assess the impact of the underlying environment representation (3DGS vs. ITG) on a state-of-the-art zero-shot navigation policy, rather than conducting a broad comparison of different navigation policies such as GNM~\cite{shah2023gnm} or ViNT~\cite{shah2023vint}. Therefore, we utilize variants of the same NoMaD policy for all methods to ensure a fair comparison of the representation itself.

\subsubsection{Configurations}
For all methods, including the proposed GSplatVNM, we utilize the same image DB and pre-trained NoMaD weights provided by the authors~\cite{sridhar2024nomad}. Note that the pre-trained datasets do not include data from the AI Habitat simulator.

\subsection{Experimental Results}

\subsubsection{Comparisons of the Navigation Efficiency}

\looseness=-1 
Figure~\ref{fig:spl} presents the SPL distribution over $N_{\rm{trials}}=20$ trials for each environment and image DB size. Overall, the proposed GSplatVNM achieves higher SPL values compared to the baseline methods across all environments and image DB sizes.

We observe that the baseline methods exhibit varying trends depending on the image DB size. For NoMaD w/o ITG, the SPL is generally lower than that of the other methods, particularly when the image DB is sufficiently large, because it merely traverses the pre-collected image sequence, leading to frequent detours. Interestingly, when the image DB is very sparse, SPL can be higher than in denser cases, as an untrackable sequence of point-of-views may cause the robot to deviate from the intended path, sometimes resulting in a fortunate shortcut. In contrast, NoMaD w/ ITG shows significant degradation in SPL as the image DB size decreases—especially in the \mytag{Ribera} and \mytag{skokloster-castle} environments—due to a reduced number of nodes in the ITG, which may lead to cases where the start and goal images are absent or detours occur.

In contrast, GSplatVNM demonstrates robustness with respect to the image DB size in terms of SPL. Specifically, for image DB sizes of 500 and 1000 in \mytag{Greigsville}, 500, 800, and 1000 in \mytag{Ribera}, and 1000, 2000, and 3000 in \mytag{skokloster-castle}, GSplatVNM achieves consistently high SPL values. For very small image DBs (300 images for \mytag{Greigsville} and \mytag{skokloster-castle}), SPL decreases slightly due to a decline in the quality of images rendered by 3DGS, which adversely affects pose estimation accuracy and tracking performance of NoMaD.

\subsubsection{Qualitative Comparisons}

\begin{figure*}[t]
    \centering
    \subfloat[A simulation result on \mytag{Greigsville} (image DB size: 500)\label{fig:traj_greigsville}]{%
        \includegraphics[width=\linewidth]{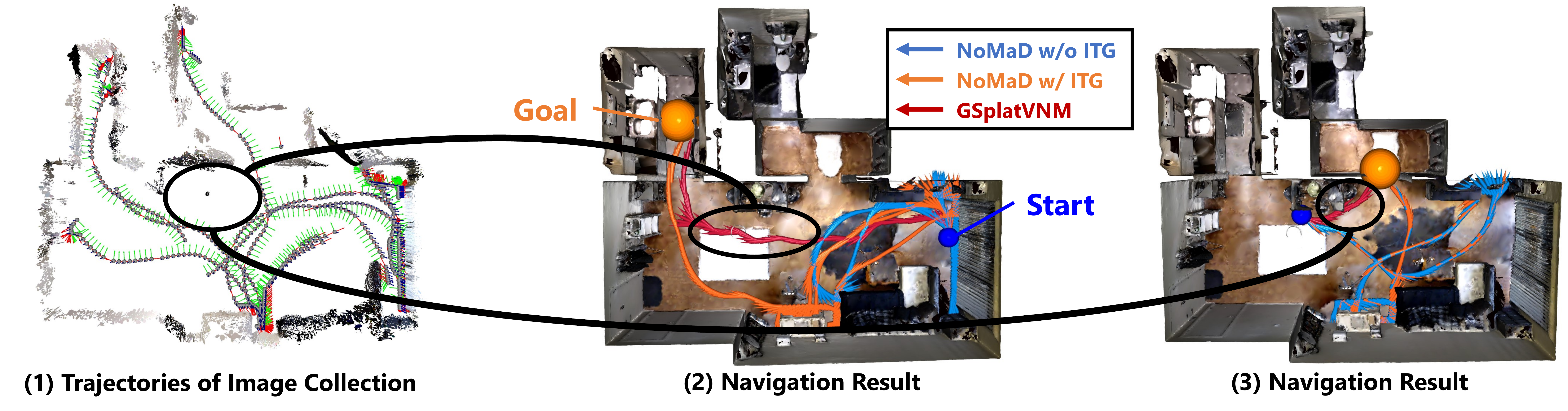}
    }\\[1mm]
    \subfloat[A simulation result on \mytag{Ribera} (image DB size: 800)\label{fig:traj_ribela}]{%
        \includegraphics[width=\linewidth]{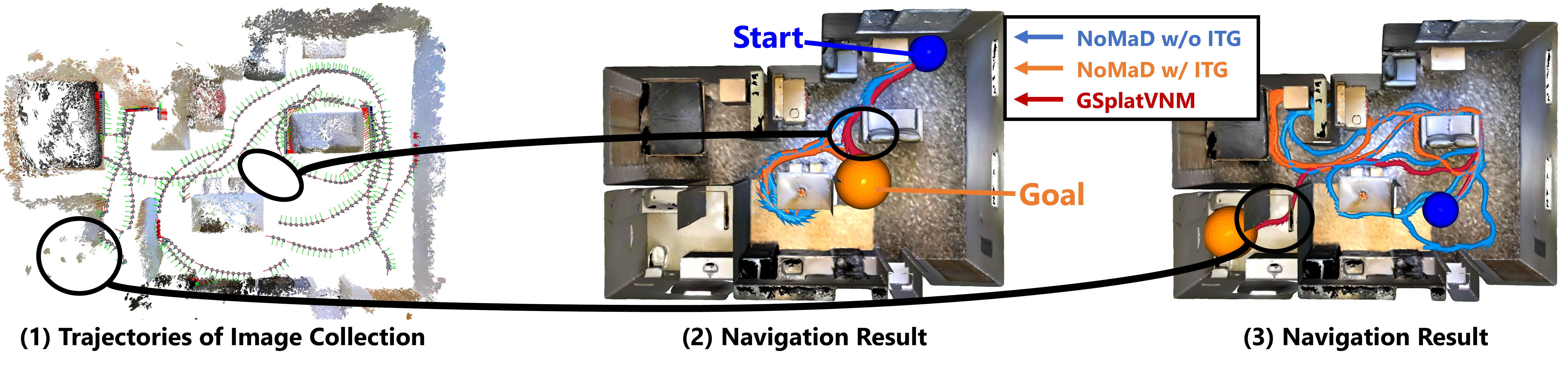}
    }\\[1mm]
    \subfloat[A simulation result on \mytag{skokloster-castle} (image DB size: 1000)\label{fig:traj_skokloster_castle}]{%
        \includegraphics[width=\linewidth]{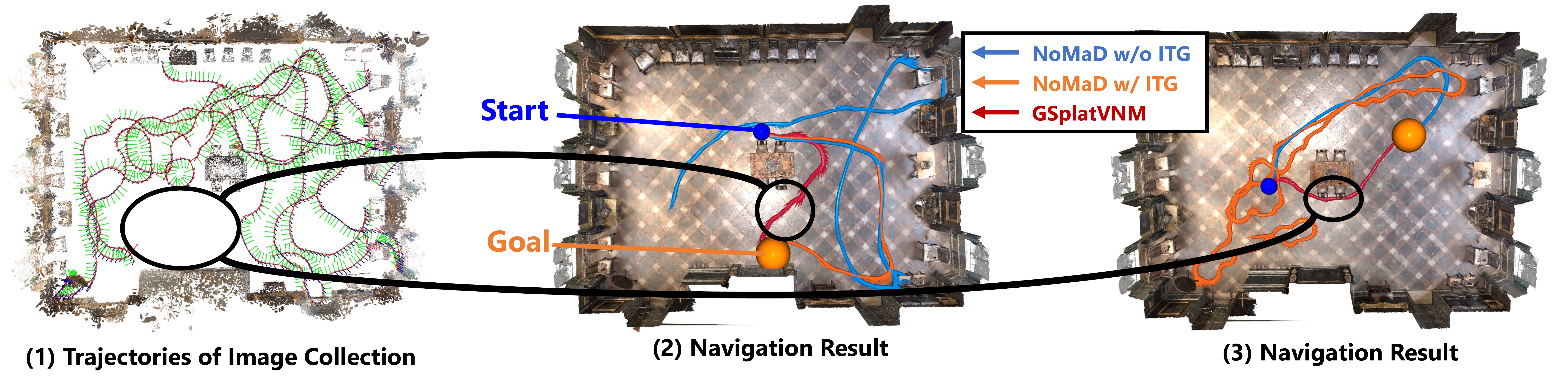}
    }
    \caption{Trajectories of the image collection and selected navigation results for each environment. GSplatVNM can generate point-of-view images that are not included in the pre-collected image DB, enabling the robot to reach the goal via a shorter path than the baseline methods. Notably, GSplatVNM is able to reach a goal located in a room that was not visited during image collection (see Fig.~\ref{fig:traj_ribela} (3)).}
    \label{fig:simulation_traj}
\end{figure*}

Figure~\ref{fig:simulation_traj} shows selected navigation results for both the baseline methods and GSplatVNM. In Figures~\ref{fig:traj_greigsville} and \ref{fig:traj_skokloster_castle}, GSplatVNM successfully interpolates point-of-view images in the gaps of the image collection trajectory, allowing the robot to reach the goal along a shorter path compared to the baselines. In Figure~\ref{fig:traj_ribela} (3), the goal is located in a bottom-left room that is inaccessible using only the image DB, as the robot did not enter that room during image collection. In this case, while the baseline methods failed, GSplatVNM succeeded by synthesizing the missing point-of-view images for areas that were observed but not visited.

\subsubsection{Performance Analysis of Individual Modules}

\begin{table}[t]
    \centering
    \caption{Success Rate of the Navigation Task and Individual Modules}
    \label{tab:benchmark_results}
    \begin{tabular}{llccc}
      \toprule
      Env. ID & Method & SR  & PE-SR  & VF-SR \\
      \midrule
      \multirow{3}{*}{\scriptsize{\mytag{Greigsville}}}
             & NoMaD w/o ITG & 0.47 & 0.27 & 0.22 \\
             & NoMaD w/ ITG  & 0.42 & 0.27 & \textbf{0.48} \\
             & GSplatVNM    & \textbf{0.55} & \textbf{0.39} & 0.38 \\
      \midrule
      \multirow{3}{*}{\scriptsize{\mytag{Ribera}}} 
             & NoMaD w/o ITG & 0.30 & 0.34 & 0.05  \\
             & NoMaD w/ ITG  & 0.27 & 0.34 & \textbf{0.51} \\
             & GSplatVNM    & \textbf{0.40} & \textbf{0.47} & 0.33 \\
      \midrule
      \multirow{3}{*}{\scriptsize{\mytag{skokloster-castle}}}
             & NoMaD w/o ITG & 0.34 & 0.20 & 0.18 \\
             & NoMaD w/ ITG  & 0.39 & 0.20 & \textbf{0.59} \\
             & GSplatVNM    & \textbf{0.59} & \textbf{0.62} & 0.37 \\
      \bottomrule
    \end{tabular}
\end{table}

Table~\ref{tab:benchmark_results} summarizes the overall navigation success rate (SR) along with the success rates of individual modules: the Pose Estimation Success Rate (PE-SR) for the start and goal poses, and the target point-of-view Following Success Rate (VF-SR). The PE-SR reflects the accuracy of the estimated start and goal poses (manually verified), while the VF-SR measures whether NoMaD successfully reaches the final node in the sequence of target point-of-view images, indicating the ease of following by NoMaD.

Across all environments, GSplatVNM consistently achieves the highest SR, with particularly notable performance in the \mytag{skokloster-castle} environment, where its SR is 20\% higher than that of NoMaD w/ ITG. Additionally, GSplatVNM exhibits a higher PE-SR than the baseline methods, demonstrating its ability to optimize point-of-view images despite lower image similarity on the 3DGS. Notably, in the \mytag{skokloster-castle} environment, where the image DB has limited spatial coverage, GSplatVNM achieves a high PE-SR by effectively interpolating point-of-view images. However, regarding VF-SR, NoMaD w/ ITG attains the highest success rates in the \mytag{Ribera} and \mytag{skokloster-castle} environments. This is because the point-of-view images generated by ITG are closer to the actual traversed views, facilitating easier tracking by NoMaD. Improving NoMaD’s tracking performance along these generated views represents a promising direction for future research.

\subsubsection{Storage Usage Comparison}

\begin{table}[t]
    \centering
    \caption{Comparison of Storage Usage (skokloster-castle)}
    \label{tab:storage_comparison}
    \begin{tabular}{lcc}
      \toprule
      \textbf{Image Num.} & \textbf{ITG Storage (MB)} & \textbf{3DGS Storage (MB)} \\
      \midrule
      300   & 322  & 508  \\
      1000  & 840  & 530  \\
      2000  & 2300 & 442  \\
      3000  & 3400 & 223  \\
      \bottomrule
    \end{tabular}
\end{table}

Table~\ref{tab:storage_comparison} compares the storage usage of ITG and 3DGS for various image DB sizes in the \mytag{skokloster-castle} environment. The ITG requires significantly more storage than 3DGS, with its storage usage increasing linearly with the number of images in the DB. In contrast, the storage usage of 3DGS is independent of the number of images, as it compresses image information into a compact representation based on Gaussians; thus, the storage depends solely on the number of Gaussians. These results confirm that 3DGS is a more storage-efficient representation for VNMs.

\section{CONCLUSION}

\looseness=-1 
This paper proposes a novel environment representation for VNMs using 3DGS instead of the traditional ITG. With 3DGS, spatially arbitrary point-of-view images can be generated between the start and the goal, enabling high navigation performance even when ITG lacks sufficient spatial coverage. As a result, the robot is capable of reaching locations that \emph{have been observed but not previously visited}. Furthermore, 3DGS offers improved storage efficiency compared to ITG, which is particularly advantageous for robotics applications.

Despite these promising results, our work has several limitations:
\begin{itemize}
\item \textbf{Real-World Validation}: The current validation is confined to simulation environments. A crucial next step is to conduct experiments on a physical robot to evaluate the sim-to-real transferability and robustness of our framework under real-world conditions.
\item \textbf{Planar Motion Assumption}: Our framework assumes the robot operates on a 2D plane, making it unsuitable for environments with significant elevation changes, such as ramps or multiple floors. While our approach could theoretically be extended by incorporating 3D pose estimation and global planning, we leave this extension for future work.
\item \textbf{Collision Checking with 3DGS}: The collision checking in our method lacks geometric precision. 3DGS is optimized for rendering quality rather than metrically accurate reconstruction, which can cause a scale discrepancy between the model and the physical world. Consequently, the robot's radius for collision checking within the 3DGS space is treated as a hyperparameter in our experiment. However, this limitation is mitigated by the fact that the VNM policy follows rendered target images instead of precise world coordinates, reducing the criticality of exact geometric collision checking.
\end{itemize}

Looking forward, several technical challenges remain to be addressed. One is improving the tracking performance of the VNM on the generated point-of-view images. Potential approaches include fine-tuning the VNM itself~\cite{shah2023vint} or re-planning the global trajectory~\cite{honda2024replan}. Another challenge is handling dynamic obstacles. Although the current VNM can avoid obstacles, online reconstruction of 3DGS for dynamic environments~\cite{yan2024street} could provide a more robust solution. Finally, enhancing the global position estimation accuracy within the 3DGS representation is important. Since image similarity-based methods can sometimes confuse visually similar locations, incorporating instance segmentation~\cite{lei2025gaussnav, garg2024robohop} may help overcome this limitation.




\balance


\bibliographystyle{IEEEtran}
\bibliography{IEEEabrv, reference}

\end{document}